\documentclass[10pt,twocolumn,letterpaper]{article}

\usepackage[pagenumbers]{cvpr} 
\usepackage[misc]{ifsym} 
\usepackage{bm}
\usepackage{amsfonts}
\usepackage{amsmath}
\usepackage{booktabs}
\usepackage{multirow}
\usepackage{algorithm}
\usepackage{algorithmic}
\usepackage{color}
\usepackage{xcolor}
\definecolor{Red}{RGB}{255,0,0}
\definecolor{Orange}{RGB}{255,165,0}
\definecolor{BlueGreen}{RGB}{0,169,154}
\definecolor{OliveGreen}{RGB}{128,128,0}

\definecolor{cvprblue}{rgb}{0.21,0.49,0.74}
\usepackage[pagebackref,breaklinks,colorlinks,allcolors=cvprblue]{hyperref}

\def\paperID{13506} 
\def\confName{CVPR}
\def\confYear{2025}

\title{SAM-REF: Introducing Image-Prompt Synergy during Interaction for Detail Enhancement in the Segment Anything Model}

\author{
   Chongkai Yu\textsuperscript{\rm 1,\dag},
   Ting Liu\textsuperscript{\rm 1,\dag,\Letter},
   Anqi Li\textsuperscript{\rm 2},
   Xiaochao Qu\textsuperscript{\rm 1},
   Chengjing Wu\textsuperscript{\rm 1},
   Luoqi Liu\textsuperscript{\rm 1},
   Xiaolin Hu\textsuperscript{\rm 3,\Letter}
\\ 
\small \textsuperscript{\rm 1} MT Lab, Meitu Inc., Beijing 100083, China \\
\small \textsuperscript{\rm 2} Beijing Institute of Technology \\
\small \textsuperscript{\rm 3}Department of Computer Science and Technology\\
\small BNRist, IDG/McGovern Institute for Brain Research\\
\small Tsinghua University, Beijing 100084, China\\
\small \texttt{\{yck, lt, qxc, ethan, llq5\}@meitu.com} \\
\small \texttt{xlhu@tsinghua.edu.cn} \\
\vspace{1mm}
\small \textsuperscript{$\dagger$}Joint first authors. \small \hspace{3mm} \textsuperscript{\Letter\ }Joint corresponding author.
\vspace{-0.75cm} 
}

\begin{document}
\maketitle
\begin{abstract} 
Interactive segmentation is to segment the mask of the target object according to the user's interactive prompts.
There are two mainstream strategies: \emph{early fusion} and \emph{late fusion}.
Current specialist models utilize the \emph{early fusion} strategy that encodes the combination of images and prompts to target the prompted objects, yet repetitive complex computations on the images result in high latency.
\emph{Late fusion} models extract image embeddings once and merge them with the prompts in later interactions. This strategy avoids redundant image feature extraction and improves efficiency significantly.
A recent milestone is the Segment Anything Model (SAM). However, this strategy limits the models' ability to extract detailed information from the prompted target zone.
To address this issue, 
we propose SAM-REF, a two-stage refinement framework that fully integrates images and prompts by using a lightweight refiner into the interaction of late fusion, 
which combines the accuracy of early fusion and maintains the efficiency of \emph{late fusion}.
Through extensive experiments, we show that our SAM-REF model outperforms the current state-of-the-art method in most metrics on segmentation quality without compromising efficiency.
\end{abstract}    
\section{Introduction}
\label{sec:intro}

\begin{figure}[t!]
\centering
\includegraphics[width=\columnwidth]{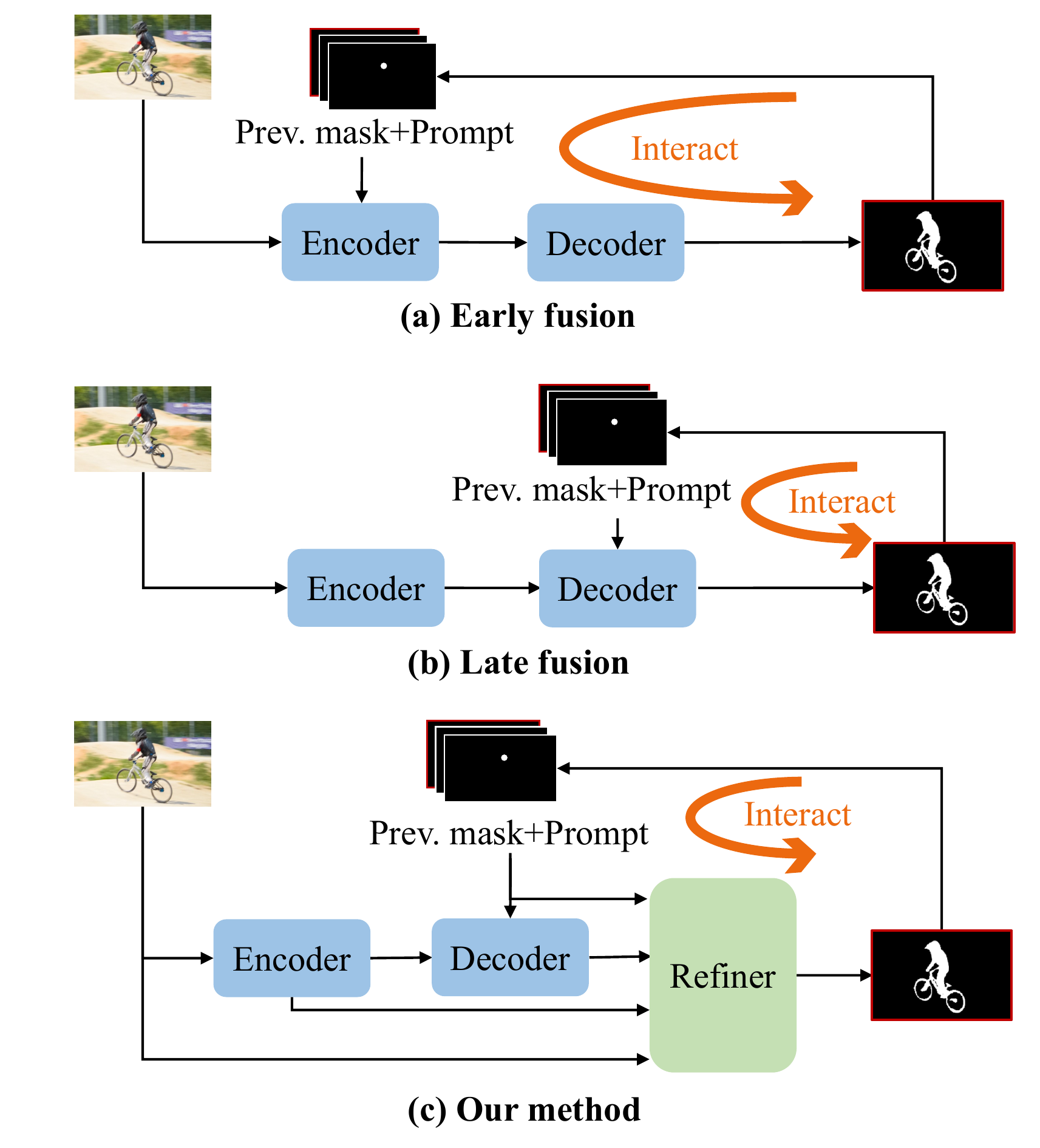} %
\caption{{\bf Three pipelines for interactive image segmentation.}
(a) Early fusion pipeline. (b) Late fusion pipeline. (c) Our proposed pipeline, which takes advantages of both early fusion and late fusion pipelines. The orange arrows indicate the interaction between the user and the systems.  }

\label{fig1}
\end{figure}

Interactive segmentation \cite{liu2024segnext,huang2024focsam,kirillov2023sam,chen2022focalclick,huang2023interformer} is an important task in computer vision that has garnered significant attention from both academia and industry. It leverages user input through simple interactions, such as scribbles, boxes, and clicks, to conveniently obtain accurate masks. 
Based on the timing of prompt-image fusion, existing methods can be categorized into early fusion strategies and late fusion strategies, as shown in Fig. \ref{fig1}.

Early fusion methods such as FocalClick \cite{chen2022focalclick}, and SimpleClick \cite{liu2023simpleclick} jointly encode the image and prompts during interaction (Fig. \ref{fig1}a). 
Since the encoder can generate features specifically for target objects, these early fusion pipelines can achieve accurate segmentation results, especially for complex and challenging cases like thin objects.
However, encoding the image in each interaction leads to low efficiency.
In contrast, late fusion methods like the Segment Anything Model (SAM) \cite{kirillov2023sam} and InterFormer \cite{huang2023interformer} extract features of the image before interaction, and fuse the prompt with the features during each interaction (Fig. \ref{fig1}b).
These methods often utilize a powerful image encoder, Vision Transformer (ViT) \cite{dosovitskiy2020vit}, to generate embeddings with rich information. 
SAM \cite{kirillov2023sam} also used a billion-scale large dataset for training to enhance the encoder's extraction capability.
Late fusion strategy avoids redundant image feature extraction, improving training and inference efficiency significantly.
However, despite the ability of a powerful image encoder to focus on all objects, it struggles to capture the fine details of the target objects, because it discards the interaction between the original image and the prompt.
This always leads to unsatisfactory results in many complex scenarios.
Subsequent finetune approaches \cite{ke2024hqsam, huang2024focsam, xie2024pa} based on SAM are also limited by the image embeddings.

In this paper, we present a method called SAM-REF which takes advantage of both early fusion and late fusion pipelines. The key idea is to utilize the late fusion pipeline for its high efficiency but integrate the prompt and the input image during the user's interaction with the system, which provides precise instructions of the user and fine details of the image.  To achieve this goal, we design a lightweight refiner and place it after SAM (Fig. \ref{fig1}c). The overall process is as follows. 
First, SAM generates the image embeddings with strong semantic representation capability and the prompt embeddings,
and then combines them into the decoder to generate the coarse mask.
Then the lightweight refiner uses a coarse-to-fine structure that can utilize global and local information.
It obtains the semantic information of the image from SAM's embeddings, and processes the details by reintroducing image-prompt synergy.
The light refiner generates a refined mask as the final output.

Experimentally, SAM-REF fully exploits the potential of SAM for interactive segmentation. On widely used benchmark datasets (GrabCut \cite{rother2004grabcut}, Berkeley \cite{mcguinness2010berkeley}, DAVIS \cite{perazzi2016davis}, and SBD \cite{hariharan2011sbd}) and high-quality benchmark dataset HQSeg-44K \cite{ke2024hqsam}, our method achieves state-of-the-art performance in most metrics while maintaining SAM's low latency advantage.

In summary, we propose SAM-REF, a new pipeline 
that leverages the power of SAM but integrates the prompt and the image during the interaction process to capture object details, leading to both high accuracy and high efficiency.

\section{Related Work}
In this paper, we focus on interactive segmentation. There are mainly two methods, early fusion and late fusion interactive segmentation method, divided based on the timing of prompt-image fusion.
\subsection{Early Fusion Methods}
Early fusion interactive segmentation \cite{barnum2020earlyfusion} means combining different information in the early stages of networks. 
It combines the image with interactive information before processing them through deeper network layers, \eg ViT \cite{dosovitskiy2020vit}, which is capable of producing more refined and contextually aware segmentation results. 
The first deep-learning-based interactive segmentation model, DIOS \cite{xu2016dios}, takes the early fusion strategy by embedding positive and negative clicks into distance maps and concatenating them with the original image as input. 
RITM \cite{sofiiuk2022ritm} introduces concatenating the previous mask with point maps, and later works mostly follow this approach. 
Many methods have been used to improve accuracy.
For example, FocalClick \cite{chen2022focalclick} and FocusCut \cite{lin2022focuscut} use local refinement to optimize the details.
CDNet \cite{chen2021cdnet} proposes a pixel diffusion module to fully fuse the prompt and the image.
SimpleClick \cite{liu2023simpleclick} up-samples the image feature, and PseudoClick \cite{liu2022pseudoclick} simulates an additional prompt.
Though high precision, encoding the image and prompts every interaction leads to the high latency of early fusion, which has always been a problem. Some work has tried to reduce latency. FocalClick \cite{chen2022focalclick} compresses the size of the feature to speed up, but this method loses some accuracy.

\begin{figure*}[th!]
\centering
\includegraphics[width=\textwidth]{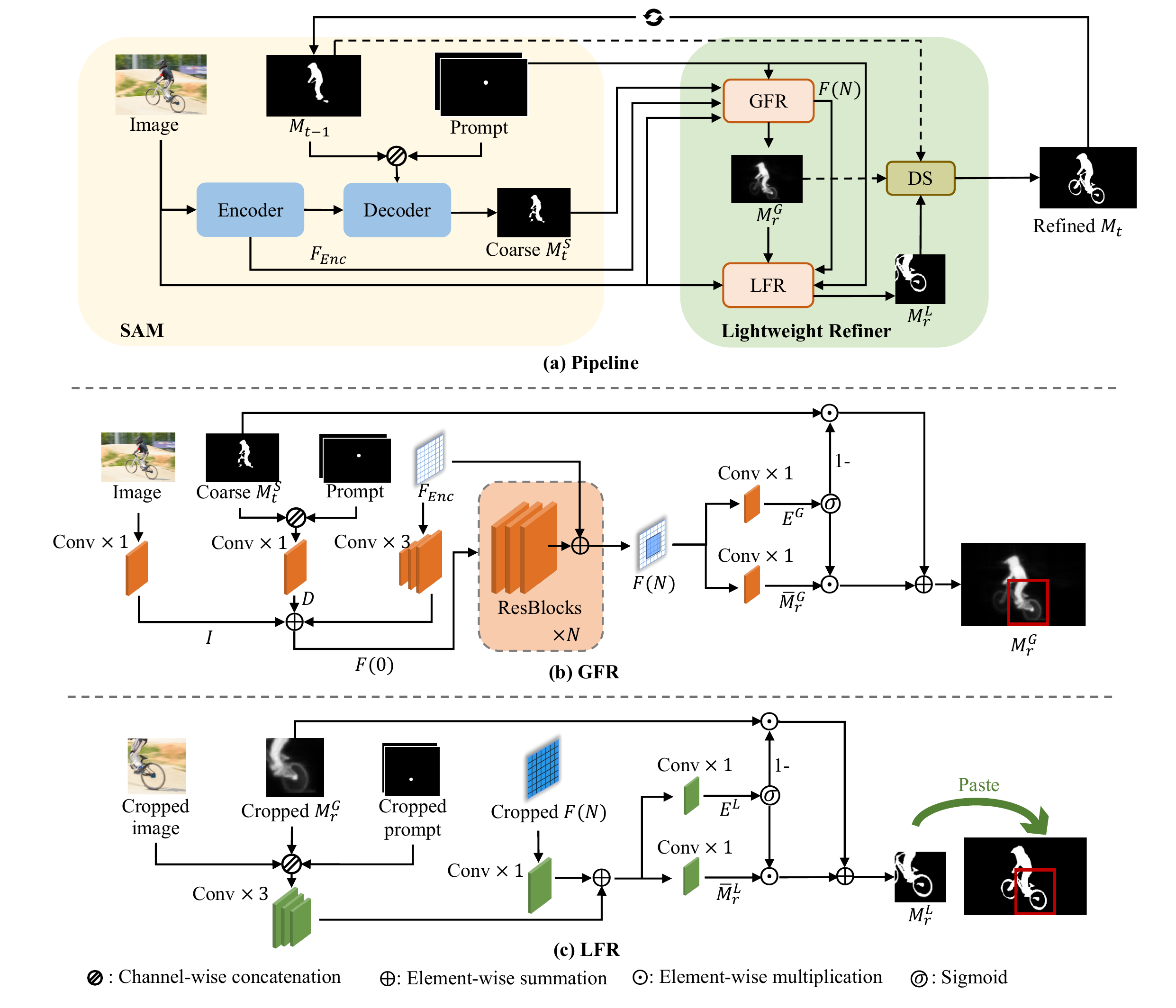} 
\caption{{\bf The framework of the proposed SAM-REF.}
We propose a decoupled lightweight refiner while maintaining the inherent structure of SAM \cite{kirillov2023sam} (a).
The refiner mainly has three modules, the Global Fusion Refiner (GFR), the Local Fusion Refiner (LFR), and the Dynamic Selector (DS).
In GFR (b) and LFR (c), Conv $\times n$ represents an n-layer convolutional network. }

\label{fig2}
\end{figure*}

\subsection{Late Fusion Methods}
Late fusion strategy extracts image embeddings once and combines them with interactive prompts in later interactions. Interformer \cite{huang2023interformer} and SAM \cite{kirillov2023sam} decouple the light prompt encoder and the decoder from the heavy encoder, accelerating the speed. However, late fusion cannot catch details according to the focusless embeddings, leading to low precision. To promote accuracy, SegNext \cite{liu2024segnext} adds dense representation and fusion of visual prompts to extract more visual information, FocSAM \cite{huang2024focsam} introduces window-based attention to focus on the target zone of embeddings, while HQ-SAM \cite{ke2024hqsam} uses global-local fusion and proposed HQ-SAM dataset with high quality. These methods aim to explore more information from the embeddings. 
However, the embeddings may not have enough detailed information about the target zone, so these methods cannot achieve significant improvements in accuracy. In contrast, our method combines early fusion and late fusion, extracting detailed information from the image directly.

\section{Method}
Since our proposed method is built upon SAM, we first briefly review the architecture of SAM. 
Then we describe our method SAM-REF, which has three major components, Global Fusion Refiner (GFR), Local Fusion Refiner (LFR), and a Dynamic Selector (DS).
Finally, we introduce the training loss.

\subsection{Preliminaries: SAM}
SAM consists of three modules: an image encode, a prompt encode, and a mask decode. \textbf{Image encoder}: It is a heavy ViT-based backbone, structured into four stages with equal depth. Each stage employs window-based attention to enhance computational efficiency \cite{li2022exploring}.
\textbf{Prompt encoder}: 
It is a lightweight structure consisting of MLP and CNN.
In the interaction phase, it transforms the positional information of the input boxes, clicks, and masks into embeddings.
However, the sparse representations of visual prompts, similar to those used in linguistic prompts, may inhibit the ability to capture image details \cite{liu2024segnext}.
\textbf{Mask decoder}:
It is a lightweight structure composed of a Transformer \cite{vaswani2017attentiontransformer} decoder and a convolutional mask head.
It combines image embeddings and prompt embeddings through a two-layer transformer-based decoder to predict the final masks.

During the interaction process, the prompt encoder and mask decoder are lightweight, and the heavy image encoder only needs to be performed once before the interaction. Therefore, the entire inference pipeline is very efficient.

\subsection{Our Pipeline of the SAM-REF}
To preserve the inherent structure of SAM while fully exploiting its potential, we adopt a decoupled design approach wherever possible. To this end, we introduce two modules GFR and LFR to refine the mask progressively. 

The overall architecture of SAM-REF is shown in Fig. \ref{fig2}.
First, we use a late fusion approach to combine the encoded image with the prompt, producing a coarse mask, just like SAM \cite{kirillov2023sam}. 
Then we use the lightweight refiner, a decoupled module from SAM, to refine the coarse mask in a coarse-to-fine manner. 
In addition to the coarse mask and the prompt, the image and the feature from the encoder are also input to the lightweight refiner. 
Inside the lightweight refiner, we first use the global fusion refiner (GFR) to directly combine the prompt with the image and extract detailed information about the prompted target with a lightweight convolutional network.
Meanwhile, it progressively incorporates SAM's image embeddings at each stage to preserve SAM's robust deep semantic capabilities, thus refining the overall objects.

To further refine the local details of the target region, we propose the local fusion refiner (LFR). It crops the target region of the mask from GFR according to the prompt, refines the cropped mask, and pastes it back to the mask.
We propose a dynamic selector to adaptively determine which mask to paste on.
The cropped mask can be pasted on the predicted result of GFR or the previous mask as the output. The former can re-predict the full mask, while the latter can only refine the target local region without modifying other regions.

\subsection{Global Fusion Refiner (GFR)}
\subsubsection{Fusion of Image, Prompt, Coarse Mask and Encoder Features}
As stated in Sec. \ref{sec:intro}, we need to efficiently fuse the prompt and the image. 
Unlike SAM \cite{kirillov2023sam} that represents the prompts with sparse one-dimension vectors, we use a dense presentation of prompts (disks with a small radius) to match the spatial dimensions of the image \cite{liu2024segnext, sofiiuk2022ritm, benenson2019large, chen2022focalclick}, which better preserving the detailed spatial attributes of the visual prompts.
We encode the prompts and the coarse mask $\bm{M}_{t}^S$ ($S$ refers to SAM, and $t$ refers to the $t$-th interaction) to generate a dense map $D$.
The dense map is the concatenation of the positive prompt map, the negative prompt map, and the coarse mask $\bm{M}_{t}^S$. 
We adopt the method from RITM \cite{sofiiuk2022ritm}, utilizing two convolutions to align the scale and channels of the image 
and the dense map from 
$\bm{I}_{ori}\in\mathbb{R}^{3\times 1024\times 1024}$, $\bm{D}_{ori}\in\mathbb{R}^{3\times 256\times 256}$ to $\bm{I}\in\mathbb{R}^{64\times 256\times 256}$, $\bm{D}\in\mathbb{R}^{64\times 256\times 256}$.

We assume that SAM, relying on the powerful ViT architecture, has already learned deep semantic information \cite{huang2024segment}. 
The primary deficiency lies in its understanding of low-level details. 
To maintain SAM's semantic understanding and design an efficient detail extraction module, we employ a scale-invariant convolutional structure and integrate it with SAM image embeddings $\bm{F}_{Enc}\in\mathbb{R}^{256\times 64\times 64}$.

Specifically, as illustrated in Fig. \ref{fig2}b, we progressively upscale the SAM image embeddings $\bm{F}_{Enc}$ to $\bm{\hat{F}}_{Enc}\in\mathbb{R}^{64\times 256\times 256}$.
Then we fuse the image, prompts, coarse mask, and encoder features by adding them together:
\begin{equation}
\label{eq:f0}
    \bm{F}\left(0\right)=\bm{I}+\bm{D}+\bm{\hat{F}}_{Enc}.
\end{equation}
Here $\bm{D}$ contains the information of prompts and the coarse mask. 
Subsequently, we utilize a scale-invariant ResBasicBlock \cite{he2016deep} to capture fine details. This process is repeated $N$ times, with each iteration merging with the SAM image embeddings to prevent the loss of semantic information from SAM:
\begin{equation}
\label{eq:fn}
    \bm{F}\left(i+1\right)=ResBlock\left(\bm{F}\left(i\right)\right)+\bm{\hat{F}}_{Enc}, i\in[0,N-1].
\end{equation}
$\bm{F}(N)$ contains the information of image, prompt, coarse mask, and encoder features.

\subsubsection{Fusion of Coarse Mask and Refined Mask}
Since the coarse mask $\bm{M}_{t}^S$ is the prediction of SAM, it is generally accurate. We want to identify the inaccurate areas of $\bm{M}_{t}^S$, and only modify these areas.
GFR predicts a mask result $\Bar{\bm{M}}_r^G$ and an error map $\bm{E}^G$, which represents error-prone regions. We replace these error-prone regions in $\bm{M}_{t}^S$ with the corresponding regions in $\bm{M}^G_r$. The fusion process is formulated as follows.
\begin{equation}
    \bm{M}_{r}^{G}=~\sigma\left(\bm{E}^{G}\right){\odot}\Bar{\bm{M}}_{r}^{G}{+}\left(1{-}\sigma\left(\bm{E}^{G}\right)\right){\odot}\bm{M}_{t}^S,
\label{eq:mgr}
\end{equation}
where $\sigma(\cdot)$ represents Sigmoid function. $\bm{E}^{G}$ is supervised by the error regions between the ground truth and the binary mask.
After GFR, we obtain the global refinement of the objects. 

\subsection{Local Fusion Refiner (LFR)}
To further refine the local details of the target region, we propose LFR to refine the local region in $\bm{M}_r^G$.
In this module, we perform prompt and image fusion again for the user-interacted local regions to refine the local details of the objects. Inspired by FocalClick \cite{chen2022focalclick}, we calculate the max connected component of the difference between the GFR result and the previous mask. Subsequently, we generate the external box for the connected component that includes the last prompt and proportionally expand it. 
The specific design of the LFR is illustrated in Fig. \ref{fig2}c.

In contrast to FocalClick, we use RoiAlign \cite{he2017mask} to crop the logits and the fused feature $\bm{F}\left(N\right)$ from GFR. 
Similar to GFR,
our two prediction heads generate the local mask result $\Bar{\bm{M}}_r^L$ and the error map $\bm{E}^L$ based on the prediction of GFR $\bm{M}_r^G$.
The local refined mask $M_{r}^{L}$ fuses $\Bar{\bm{M}}_r^L$ and $\bm{M}_r^G$ as follows:
\begin{equation}
\bm{M}_{r}^{L}=~\sigma \left(\bm{E}^{L}\right){\odot}\Bar{\bm{M}}_{r}^{L}{+}\left(1{-}\sigma\left(\bm{E}^{L}\right)\right){\odot}\bm{M}_{r}^{G}.
\label{eq2}
\end{equation}

\subsection{Dynamic Selector}
As stated in FocalClick \cite{chen2022focalclick}, when the previous mask $\bm{M}_{t-1}$ is already well-refined, especially when it has been refined for many rounds, re-predicting the full mask 
may destroy areas that have already been well refined.
FocalClick \cite{chen2022focalclick} always pastes the refined local mask onto the previous mask after $10$ interactions,
as by this point the previous mask typically converges to good segmentation quality.
In this paper, we use the dynamic selector to select which mask should the local result $\bm{M}_r^L$ be pasted onto, the global refined mask $\bm{M}_r^G$ or the previous mask $\bm{M}_{t-1}$.
Taking the global result $\bm{M}_r^G$ pasted with $\bm{M}_r^L$ as the output
can refine the local area based on the refined full mask, which is suitable when
re-predicting the full mask will lead to less error.
And when re-predicting the full mask leads to more error, which often happens when $\bm{M}_{t-1}$ is already precise enough,
we paste $\bm{M}_r^L$ onto $M_{t-1}$ to only refine the target local area and avoid overwriting other areas.

We adaptively decide which mask to paste onto by comparing the changes in error maps $\bm{E}^G$ between each iteration of GFR. 
We use the error ratio $R_e$, formulated as:
\begin{equation}
    R_e=\Vert H(\bm{E}^G_{t})\Vert_0~/~\Vert H(\bm{E}^G_{t-1})\Vert_0.
\end{equation}
$H(\cdot)$ returns $1$ when the value is greater than $0$, else returns $0$.
If the error ratio $R_e$ is larger than a threshold $\theta$, 
it indicates that the modification on the full mask will lead to more error,
and we choose to
paste $\bm{M}_r^L$ onto $\bm{M}_{t-1}$ as the output $\bm{M}_t$. Otherwise, paste it onto $\bm{M}_r^G$ as $\bm{M}_t$. The output $\bm{M}_t$ will become a previous mask $\bm{M}_{t-1}$ in next interaction.
In our experiment, the threshold $\theta$ is set to be $1.1$.

\subsection{Training supervision}
First, we adopted Normalized Focal Loss $\bm{L}_{nfl}$ proposed in RITM \cite{sofiiuk2022ritm} to finetune the SAM decoder. Then, we used Dice Loss $\bm{L}_{dice}$ and Binary Cross Entropy Loss $\bm{L}_{bce}$ to supervise the error heads of GFR and LFR. 
$\bm{L}_{wnfl}$ denotes weighted NFL loss for the refined predictions. 
We calculate the error regions between the ground truth and the binary mask (for GFR it is $\bm{M}_{t}^S$, and for LFR it is $\bm{M}_r^G$). We calculate the NFL loss and add weight (1.5 in our experiment) for these error regions to get $\bm{L}_{wnfl}$.
\begin{equation}
\begin{aligned}
    \bm{L}=\bm{L}_{nfl} &{+} \bm{L}_{dice}^{G} {+} \bm{L}_{bce}^{G} {+} \bm{L}_{wnfl}^{G} \\
&{+} \bm{L}_{dice}^{L} {+} \bm{L}_{bce}^{L} {+} \bm{L}_{wnfl}^{L}
\end{aligned}
\label{eq3}
\end{equation}
where the superscripts $G$ and $L$ indicate
the supervision for GFR and LFR respectively.

\section{Experiments}
\begin{table*}[t]
\centering
\footnotesize
\fontsize{8.3pt}{10pt}
\setlength{\tabcolsep}{3.5pt}
\selectfont{}
\begin{tabular}{llllcccccccc}
\toprule
\multirow{2}{*}{Method}&\multirow{2}{*}{Train data}&\multirow{2}{*}{Backbone}&\multirow{2}{*}{SPC/s$\downarrow$}&\multicolumn{2}{c}{Grabcut \cite{rother2004grabcut}}&\multicolumn{2}{c}{Berkeley \cite{mcguinness2010berkeley}}&\multicolumn{2}{c}{DAVIS \cite{perazzi2016davis}}&\multicolumn{2}{c}{SBD \cite{hariharan2011sbd}} \\
\cmidrule(lr){5-6}
\cmidrule(lr){7-8}
\cmidrule(lr){9-10}
\cmidrule(lr){11-12}
&&&&NoC90$\downarrow$&NoC95$\downarrow$&NoC90$\downarrow$&NoC95$\downarrow$&NoC90$\downarrow$&NoC95$\downarrow$&NoC90$\downarrow$&NoC95$\downarrow$\\
\hline
\emph{Early Fusion} &&&&&&&\\
F-BRS \cite{sofiiuk2020fbrs}\textsubscript{CVPR20}&COCO+LVIS&ResNet&-&2.72&-&4.57&-&7.41&-&7.73&-\\
CDNet \cite{chen2021cdnet}\textsubscript{ICCV21}&COCO+LVIS&ResNet&-&1.52&-&2.06&-&7.04&-&5.56&-\\
EdgeFlow \cite{hao2021edgeflow}\textsubscript{ICCVW21}&COCO+LVIS&HRNet18&-&1.72&-&2.40&-&-&-&5.77&-\\
RITM \cite{sofiiuk2022ritm}\textsubscript{ICIP22}&COCO+LVIS&HRNet32&-&1.56&2.48&2.10&5.41&5.34&11.52&5.71&12.00\\
PseudoClick \cite{liu2022pseudoclick}\textsubscript{ECCV22}&COCO+LVIS&HRNet32&-&1.50&-&2.08&-&5.11&-&5.54&-\\
FocalClick \cite{chen2022focalclick}\textsubscript{CVPR22}&COCO+LVIS&SegFB3-S2$_{384}$&0.232&1.68&1.92&1.71&4.21&4.90&10.40&5.59&11.91\\
AdaptiveClick \cite{lin2024adaptiveclick}\textsubscript{TNNLS2024}&COCO+LVIS&ViT-H&1.69&1.38&-&1.64&-&4.60&-&4.68&-\\
SimpleClick \cite{liu2023simpleclick}\textsubscript{ICCV23}&COCO+LVIS&ViT-H&8.24&1.50&{\bf 1.66}&1.75&4.34&4.78&{\bf 8.88}&4.70&10.76\\ 
\hline
\emph{Late Fusion}&&&&&&&\\
InterFormer \cite{huang2023interformer}\textsubscript{ICCV23}&COCO+LVIS&ViT-B&0.153&1.50&-&3.14&-&6.19&-&6.34&-\\
InterFormer \cite{huang2023interformer}\textsubscript{ICCV23}&COCO+LVIS&ViT-L&0.271&1.36&-&2.53&-&5.21&-&5.51&-\\
SAM \cite{kirillov2023sam}\textsubscript{ICCV23}&SA-1B&ViT-H&0.413&1.88&2.28&2.09&5.14&5.19&10.00&7.62&15.03\\
HQ-SAM \cite{ke2024hqsam}\textsubscript{NeurIPS24}&HQSeg-44K&ViT-H$^*$&0.443&1.86&2.26&2.14&4.94&5.06&9.65&7.91&15.08\\
FocSAM \cite{huang2024focsam}\textsubscript{CVPR24}&COCO+LVIS&ViT-H$^*$&0.467&1.44&2.60&1.50&3.44&4.76&10.77&5.07&11.67\\
\hline
{\bf SAM-REF}&HQSeg-44K&ViT-H$^*$&0.511&{1.76}&{2.22}&{1.98}&{4.57}&{4.78}&{9.38}&{6.81}&{12.18}\\
{\bf SAM-REF}&COCO+LVIS&ViT-B$^*$&0.278&1.40&2.44&1.58&3.73& 4.75&10.45&5.56&11.76\\
{\bf SAM-REF}&COCO+LVIS&ViT-H$^*$&0.511&{\bf 1.36}&2.20&{\bf 1.43}&{\bf 3.18}&{\bf 4.54}&9.10&{\bf 4.44}&{\bf 10.41}\\
\bottomrule
\end{tabular}
\caption{{\bf Comparison of SPC, NoC90 and NoC95 with previous methods.} 
We report the results on the four mainstream datasets. 
The superscript $*$ indicates frozen backbone. 
}
\label{table:exp_result}
\end{table*}

\begin{table*}[t!]
\centering
\small
\begin{tabular}{llllcccc}
\toprule
Method&Train data&Backbone&Latency(s)$\downarrow$&5-mIoU$\uparrow$&NoC90$\downarrow$&NoC95$\downarrow$&NoF95$\downarrow$\\
\hline
FocalClick \cite{chen2022focalclick}\textsubscript{CVPR22}&COCO+LVIS&SegFB3-S2$_{256}$&30.85&84.63&8.12&12.63&835\\
FocalClick \cite{chen2022focalclick}\textsubscript{CVPR22}&COCO+LVIS&SegFB3-S2$_{384}$&40.14&85.45&7.03&10.74&649\\
SimpleClick \cite{liu2023simpleclick}\textsubscript{ICCV23}&COCO+LVIS&ViT-B&88.5&85.11&7.47&12.39&797\\
InterFormer \cite{huang2023interformer}\textsubscript{ICCV23}&COCO+LVIS&ViT-B&30.5&82.62&7.17&10.77&658\\
SegNext \cite{liu2024segnext}\textsubscript{CVPR24}&COCO+LVIS&ViT-B&22.1&85.71&7.18&11.52&700\\
\hline
SAM \cite{kirillov2023sam}\textsubscript{ICCV23}&SA-1B&ViT-B&4.0&86.16&7.46&12.42&811\\
SAM \cite{kirillov2023sam}\textsubscript{ICCV23}&SA-1B&ViT-H&4.21 &88.0 &6.50 &10.53 &653 \\
HQ-SAM \cite{ke2024hqsam}\textsubscript{NeurIPS24}&HQSeg-44K&ViT-B$^*$&4.7&89.45&6.49&10.79&671\\
FocSAM \cite{huang2024focsam}\textsubscript{CVPR24}&COCO+LVIS&ViT-H$^*$&18.6&88.6&5.74&9.44&580\\
\hline
{\bf SAM-REF}&HQSeg-44K&ViT-B$^*$&5.1&{\bf 90.75}&{5.90}&{10.08}&{607}\\
{\bf SAM-REF}&COCO+LVIS&ViT-B$^*$&5.1&88.5&6.19&9.84&603\\
{\bf SAM-REF}&COCO+LVIS&ViT-H$^*$&5.22&89.6&{\bf 5.44}&{\bf 9.16}&{\bf 566}\\
\bottomrule
\end{tabular}
\caption{{\bf Results on high-quality datasets.} All of the models are evaluated on the HQSeg-44K validation set \cite{ke2024hqsam}. 
SAM-REF achieves better results than SAM-based and other methods.
The superscript $*$ indicates frozen backbone. 
}
\label{table:hq_result}
\end{table*}

We conduct a series of experiments to test the performance of our model. We contrast our method with previous works. The experimental settings and the results of the experiments are as follows.

\subsection{Experimental Setting}
\label{exp_setting}

\subsubsection{Dataset.} We train our model on $2$ different dataset configurations. 
For one configuration we train our model on COCO \cite{lin2014coco} $+$ LVIS \cite{gupta2019lvis} datasets, following previous works \cite{huang2024focsam,chen2022focalclick,huang2023interformer,sofiiuk2022ritm, liu2023simpleclick}. 
For another configuration, we train our model on HQSeg-44K to test the performance of our method on high-precision datasets \cite{ke2024hqsam,liu2024segnext}. 
HQSeg-44K dataset is a high-quality and well-annotated dataset with $44K$ extremely accurate image mask annotations. 
 We evaluate the models' zero-shot interactive segmentation capabilities on GrabCut \cite{rother2004grabcut}, Berkeley \cite{mcguinness2010berkeley}, DAVIS \cite{perazzi2016davis}, SBD \cite{hariharan2011sbd}, and HQSeg-44K validation set, aligning with previous studies.

\subsubsection{Implementation details.}
We utilize the pre-trained ViT-H and ViT-B from SAM \cite{kirillov2023sam} as the backbone with the prompt encoder and decoder.
The GlobalFusion Refiner and the LocalFusion Refiner are light-weighted convolutional neural networks, as shown in Fig. \ref{fig2}. We use $N=3$ ResBasicBlocks to achieve the best results.
Training and evaluations are performed on a server with $4$ NVIDIA Tesla V100-PCIE-32GB GPUs and Intel(R) Xeon(R) Gold 6278C CPU.

\subsubsection{Training strategy.}
In the training process, we adopt the two-stage training strategy of FocSAM \cite{huang2024focsam} that firstly fine-tune the SAM decoder for 320k iterations at a batch size of $4$ with frozen image encoder and prompt encoder, and then train SAM-REF with the frozen decoder and frozen encoders in the same settings for 80k iterations. 
We adopt InterFormer’s click simulation strategy \cite{huang2023interformer} for interactive simulation before loss computation.
Moreover, we use the image encoder to pre-extract and store the image embeddings of the datasets to reduce computational costs.
This strategy greatly improves the training speed.

\subsubsection{Evaluation.}
We report the results on metrics of mean Intersection over Union (mIoU), number of clicks (NoC), number of failure cases (NoF), and the Seconds
Per Click (SPC) metrics, following previous work \cite{huang2024focsam, liu2024segnext, liu2023simpleclick, kirillov2023sam,chen2022focalclick}.
And we test the latency of Segmentation Anything Task (SAT Latency) following SegNext \cite{liu2024segnext}. {\bf mIoU} measures  the average intersection over union (IoU)
given a fixed number of consecutive interactions. We report the mIoU on $5$ clicks. 
{\bf NoC} measures the number of clicks required to achieve a target IoU. 
We report NoC90 and NoC95 to measure the segmentation ability.
NoC90 is a commonly used evaluation metric.
With the increasing demand for high-quality segmentation, we provide the stricter NoC95 to demonstrate the capability to handle the details of our method.
{\bf NoF} measures the number of failure cases that cannot achieve the target IoU in $20$ clicks. We report the failure cases on NoF95. 
{\bf SPC} is a commonly used metric to measure the segmentation speed, which represents the time consumed of each click.
SPC is tested on CPU.
Following SegNext \cite{liu2024segnext}, {\bf SAT Latency} measures the latency for the Segment Anything Task (SAT). We fix the input image size at $1024\times 1024$ and prompt the model with a grid of $16\times 16$ points. We compute the total time required to process all points by inputting one point at a time with our previous mask.
SAT Latency is tested on a single GPU.
Compared to SPC, SAT Latency calculates the overall latency but not one-prompt latency, which favors late fusion methods where the image is only encoded once.

\begin{figure*}[t!]
\centering
\includegraphics[width=\textwidth]{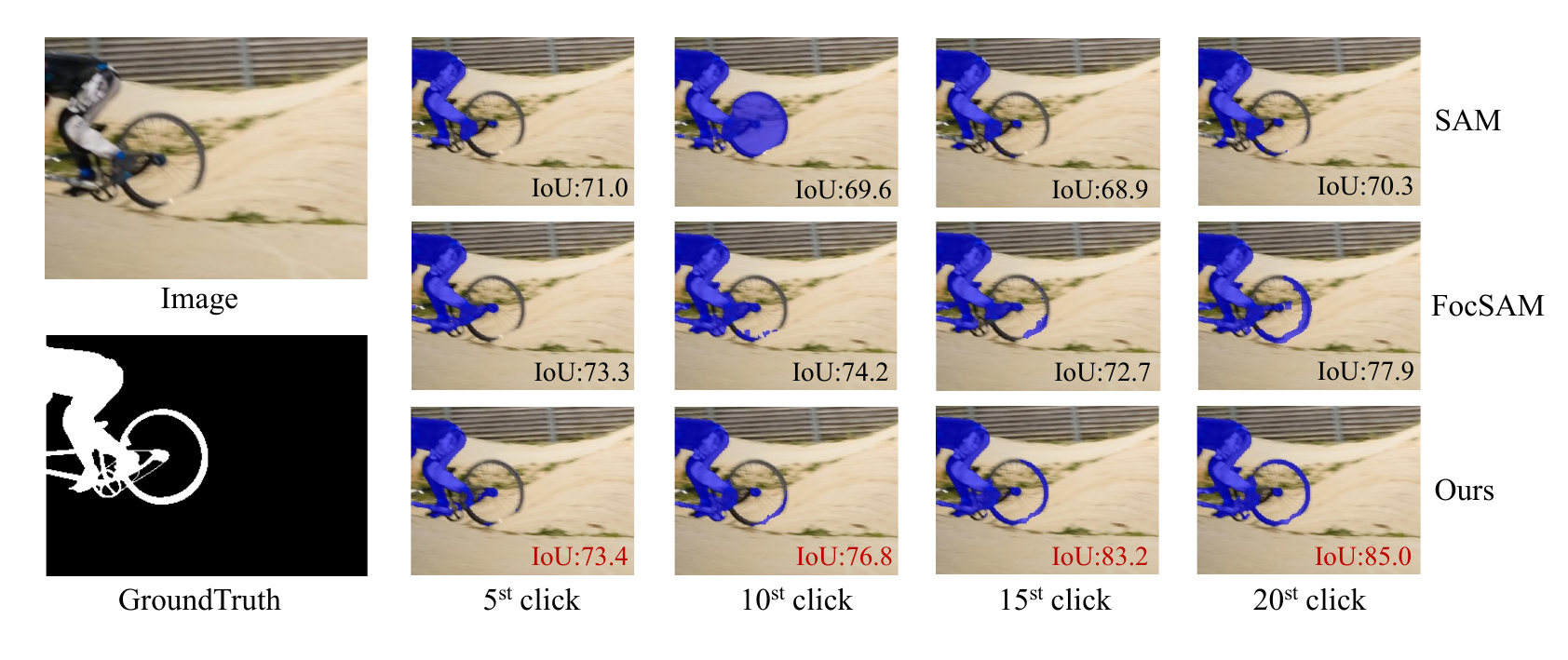} 
\caption{{\bf Qualitative analysis on a challenge Example.} The far-left image presents a challenging example featuring the image and ground truth (GT) with blue masks. The top, middle, and bottom rows on the right respectively display the segmentation results of SAM, FocSAM, and our method, at the 5th, 10th, 15th, and 20th clicks.}
\label{fig:compare_results}
\end{figure*}

\label{exp_results}
\subsection{Main Results}
{\bf Performance on widely used benchmarks.} In Tab. \ref{table:exp_result}, we present the main results of SAM-REF, benchmarked against mainstream methods. 
Our method is based on SAM, where we freeze the SAM encoder and train only a few remaining parameters to mitigate bias inherent. 
We compare our model with previous early fusion and late fusion methods. 
As reported, SAM-REF achieves state-of-the-art performance on the NoC90 in all mainstream evaluation datasets.
In addition, when applying the stricter criteria evaluation standard NoC95, our method is slightly weaker than SimpleClick only on the DAVIS and Grabcut datasets.
In the remaining datasets, SAM-REF achieves state-of-the-art performance in both NOC90 and NOC95 metrics, using ViT-H as the backbone.
However, as the early fusion method, SimpleClick requires a heavy image encoding at each interaction, whereas our approach is much more time-efficient, with only $6.2\%$ of SimpleClick's SPC. 
Additionally, we provide ViT-B results, showing it matches the speed ($0.278$ SPC) and accuracy ($4.75$ NoC90 on DAVIS) of the fastest early fusion method FocalClick \cite{chen2022focalclick} ($0.232$ SPC and $4.90$ NoC90 on DAVIS).

By reintroducing prompt-image synergy on late fusion, 
our method achieves state-of-the-art accuracy compared to all late fusion methods on these datasets. 
Specifically, under the same settings, we also outperform other SAM-based methods, such as FocSAM \cite{huang2024focsam} and HQ-SAM \cite{ke2024hqsam}.
We only add 0.1s SPC compared to SAM, which still maintain the high efficiency of late fusion.

{\bf Performance on high-quality benchmark.}
In Tab. \ref{table:hq_result}, we compare our method with state-of-the-art approaches on high-quality dataset HQSeg-44K \cite{ke2024hqsam}(1537 samples).
We compare two types of baselines: SAM-based methods and other methods.
As reported, SAM-REF achieves state-of-the-art performance to the two baselines, which demonstrate that SAM-REF can effectively enhances SAM's performance to top-tier levels in high-quality interactive segmentation while maintaining the low latency. 
Other late-fusion methods (InterFormer \cite{huang2023interformer} and SegNext \cite{liu2024segnext}), which rely solely on the dense presentation of visual prompts, are more costly than SAM's use of the sparse representation \cite{liu2024segnext}.
Our method combines the benefits of both prompt types. It only focuses on detail extraction using dense maps of visual prompts, without needing complex fusion of visual prompts and image embeddings.
This demonstrates that our method can effectively enhance the performance of late fusion in complex scenarios, as evidenced by Qualitative Results.

\begin{table}[h!]
\small
    \centering
    \begin{tabular}{cccc}
    
    \toprule
    Method&{Params/MB}&{FPS}&{Mem/G}\\
    \hline
    SAM \cite{kirillov2023sam}&{635.6}&{1.70}&{3.7}\\
    {SAM-REF}&{636.3}&{1.67}&{3.7}\\
    \bottomrule
    \end{tabular}
    \caption{\bf Computation analysis for SAM and SAM-REF}
    \label{exp_compution}
\end{table}

{\bf Computation analysis.} In Tab \ref{exp_compution}, we report the comparison of model parameters (Params), inference time per image (FPS), and GPU memory (Mem) usage between ViT-H based on SAM and SAM-REF. While SAM-REF produces substantially better segmentation quality, our method adds just 0.7MB to the model parameters compared to SAM, with negligible increases in GPU memory usage and inference time per image.

\section{Ablation Experiments}
We conduct plenty of ablation studies, reporting experimental results for ViT-H models trained on COCO+LVIS and tested on DAVIS.
\begin{table}[h!]
\centering
\begin{tabular}{lccclc}
\toprule
&{Method}&{NoC90}&{NoC95}\\
\hline
1&no $M_r^L$ only $M_r^G$&4.66&9.53\\
2&paste $M_r^L$ to $M_{t-1}$&4.80&9.75\\
3&paste $M_r^L$ to $M_r^G$&4.60&9.40\\
4&Dynamic Selector&{\bf 4.54}&{\bf 9.10}\\
\bottomrule
\end{tabular}
\caption{
{\bf Ablation study for the refiners and the  Dynamic Selector.} The methods indicate
which mask to paste 
$M_r^L$ onto.
}
\label{table:abs_refiner}
\end{table}

{\bf Component Module Analysis.}
To demonstrate the effect of our model's modules, \ie, GFR, LFR, and the Dynamic Selector, we perform an ablation study on whether and which mask to paste $\bm{M}_r^L$.
We have set up $4$ experiments. One directly outputs $M_r^G$, one always pastes $M_r^L$ onto $M_r^G$, one always pastes $M_r^L$ onto $M_{t-1}$, and one pastes $M_r^L$ according to the Dynamic Selector. For each experiment, we test NoC90 and NoC95 to evaluate the accuracy.
We have shown the results in Tab. \ref{table:abs_refiner}. 
The first experiment has no local refinement and always modifies the full mask, which leads to mutual influence between interactions and cannot achieve high accuracy.
The second experiment only refines a local region but not the full mask in each interaction, which makes it modify slowly and needs more interactions to achieve the same accuracy.
The third experiment with local refinement has a higher accuracy based on $1$-th, but still cannot solve the mutual influence.
The fourth experiment pasting $\bm{M}_r^L$ with Dynamic Selector avoids the interference and refines local regions, achieving the best accuracy.

\begin{table}[h!]
\small
    \centering
    \begin{tabular}{cccc}
    
    \toprule
    Method&SPC&NoC90&NoC95\\
    \hline
    ResBlock$_{\times1}$&0.483&4.57&9.94\\
    ResBlock$_{\times3}$&0.511&{\bf 4.54}&{\bf 9.10}\\
    ResBlock$_{\times5}$&0.545&4.66&9.88\\
    ResBlock$_{\times8}$&0.710&4.61&10.61\\
    \bottomrule
    \end{tabular}
    \caption{{\bf Ablation study for the number of ResBlocks.}}
    \label{tab:abs_cnn}
\end{table}
{\bf Number of ResBlocks.}
Tab. \ref{tab:abs_cnn} shows the impact of the number of ResBlocks on accuracy and speed. With three ResBlocks, our method achieves the best results. Fewer blocks fail to capture details effectively, while more blocks reduce speed and harm the semantic capabilities of SAM image embeddings, thus decreasing accuracy. This experiment demonstrates the rationale for setting the number of ResBlocks to $3$.

\section{Qualitative Result}

\begin{figure}[t!]
\centering
\includegraphics[width=0.45\textwidth]{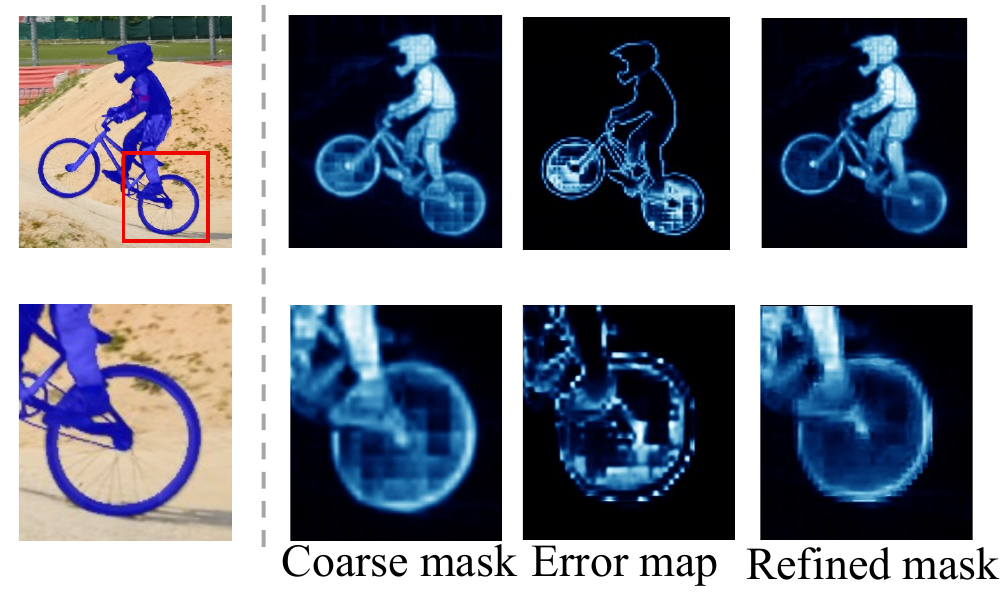} 
\caption{{\bf Qualitative results for the effectiveness of GlobalFusion and LocalFusion Refiner.} The first row shows the global refinement results of GFR. The red box indicates the region for local refinement. The second row shows the local refinement results of LFR.}
\label{fig:qualitative_results}
\end{figure}

In Fig. \ref{fig:compare_results}, we show the qualitative results compared to SAM \cite{kirillov2023sam} and FocSAM \cite{huang2024focsam} on a challenge Example.
This visualization demonstrates SAM-REF's enhanced precision. Our method achieves the highest mIoU in all clicks. 
Both FocSAM and our method are trained on COCO+LVIS and freeze the SAM's image encoder parameters.
In the later clicks, our method can segment the detailed shape of the wheel, while SAM and FocSAM cannot.
Based on challenging scenarios, the original SAM results are unsatisfactory. While FocSAM enhances stability and prevents progressive deterioration, it remains limited by image embeddings, potentially missing object details. Our method retrieves lost object details from the original image through image-prompt synergy.

We present the results at each stage of our method in Fig. \ref{fig:qualitative_results}.
The coarse mask is generated by the SAM decoder. 
The first row demonstrates how GFR refines the global segmentation process. The error map presents the error regions within SAM's predicted areas.
In this scenario, the error regions are concentrated around the object edges and thin structures.
Then, GFR specifically refines these error regions.
The second row demonstrates the refinement process of LFR. It focuses on refining the areas with the largest errors between the interactions. In this image, the errors are mainly concentrated in the hollow regions.

\section{Conclusion}
In view that early fusion strategies in the interactive segmentation can achieve high accuracy and late fusion strategies can achieve high efficiency, we propose a new method called SAM-REF, which combines the advantages of the two strategies.
Our method redesigns the SAM-based interactive segmentation pipeline, fusing images and prompts from both global and local perspectives. This adaptation fully taps into SAM's potential for interactive segmentation. SAM-REF outperforms current state-of-the-art methods while maintaining SAM's low latency, 
even in challenging scenarios.

\section{Acknowledgement}
Xiaolin Hu was supported by the National Natural Science Foundation of China (No. U2341228).

{
    \small
    \bibliographystyle{ieeenat_fullname}
    \bibliography{main}
}


\end{document}